%% file: main.tex
\documentclass[runningheads]{llncs}

\def\ECCVSubNumber{8507}
\usepackage[final,year=2026,ID=\ECCVSubNumber]{eccv}

\usepackage{graphicx}
\usepackage{amsmath,amssymb}
\usepackage{color}
\usepackage{booktabs}
\usepackage[breaklinks,colorlinks,citecolor=eccvblue]{hyperref}
\providecommand{\orcidlink}[1]{}

\makeatletter
\@ifpackageloaded{xcolor}{}{\usepackage{xcolor}}
\@ifpackageloaded{multirow}{}{\usepackage{multirow}}
\@ifpackageloaded{booktabs}{}{\usepackage{booktabs}}
\@ifpackageloaded{algorithm2e}{}{\usepackage[ruled,vlined]{algorithm2e}}
\makeatother
\input{blwritepackage}

\begin{document}

\title{Agent-OBJ: Prompt-Driven 3D Adversaries for Multi-modal Perception}
\titlerunning{Agent-OBJ: Prompt-Driven 3D Adversaries}
\author{Bing Li\inst{1} \and
        Sean Du\inst{1} \and
        Xuhong Ren\inst{2}$^{\dagger}$\and
        Luqi Gong\inst{4} \and
        Wee Peng Tay\inst{1} \and
        Qing Guo\inst{2,3}$^{\dagger}$}
\authorrunning{B. Li et al.}
\institute{Nanyang Technological University, Singapore\\
        \email{bing.li@ntu.edu.sg}
        \and
        NKIARI, Shenzhen Futian, China
        \and
        AAIS \& VCIP, Nankai University, China\\
        \email{tsingqguo@ieee.org}
        \and
        Research Center for Space Computing System, Zhejiang Lab, China}

\maketitle
\let\thefootnote\relax\footnotetext{$^{\dagger}$~ Xuhong Ren and Qing Guo are Corresponding authors.}

\begin{abstract}
Modern autonomous driving systems increasingly rely on tightly coupled Camera--LiDAR fusion pipelines to achieve robust scene understanding. While this multi-modal redundancy is designed to counter independent single-sensor failures, it implicitly assumes that physical objects cannot simultaneously deceive both modalities. In this paper, we expose the vulnerability of this assumption by proposing Agent-OBJ, a generative framework that synthesizes physically plausible 3D adversaries via a prompt-driven agent. Unlike prior works that mainly focus on sensor-specific perturbations, our pipeline instantiates a physically plausible 3D adversary by generating a canonical pedestrian geometry from a base prompt and then modulating its shape via an additional semantic prompt, while controlling appearance with a pretrained multi-view personalization model conditioned on multi-view images and a style prompt to produce view-consistent appearance. To simultaneously compromise both 2D monocular and 3D fusion detectors, the agent dynamically optimizes the generated adversary using confidence-score feedback from both detectors. Meanwhile, to ensure stealthiness, we regularize the generated 3D point cloud by enforcing similarity to a pedestrian feature bank, aligning it with the distribution of natural objects and making it visually and geometrically indistinguishable from benign instances. Furthermore, we incorporate a multi-view consistency constraint during optimization to promote cross-view robustness, ensuring the adversary remains effective under diverse viewpoints in real-world driving scenes. Extensive experiments on nuScenes demonstrate high attack success rates against both monocular and fusion detectors while preserving strong perceptual plausibility.
\end{abstract}

\begin{figure}[t] 
\centerline{\includegraphics[width=\linewidth]{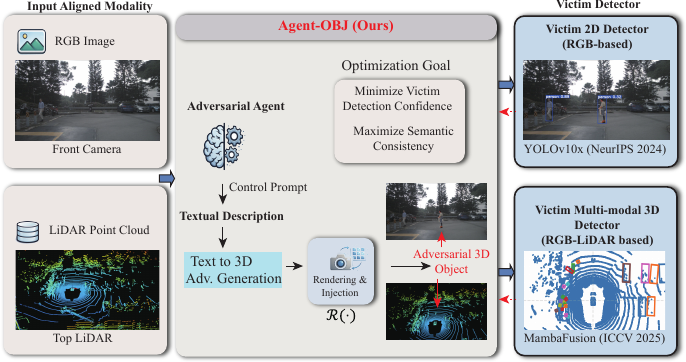}} 
\caption{Agent-OBJ pipeline. An LLM agent edits semantic and style prompts to steer a text-to-3D generator, producing a physically plausible pedestrian-like object. The object is rendered into aligned camera–LiDAR views and evaluated by both a monocular 2D detector and a Camera–LiDAR fusion 3D detector using only black-box confidence feedback.} 
\label{fig:1} 
\end{figure}

\section{Introduction}
\label{sec:intro}
Autonomous driving perception~\cite{huang2202multi,matos2024survey,wang2025mambafusion,liang2022bevfusion,shi2026intelligent} increasingly relies on multi-sensor stacks, where monocular camera perception and Camera--LiDAR fusion are widely adopted to enhance robustness. By combining complementary sensing principles, fusion systems are expected to mitigate modality-specific noise and reduce the likelihood of catastrophic failures. A common implicit safety assumption behind such redundancy is that a single \emph{physical} object is unlikely to simultaneously mislead multiple modalities in a consistent manner. In this work, we revisit this assumption and ask a fundamental question: can \emph{object-level}, physically realizable adversaries remain effective across both monocular and fusion pipelines?

Realizing such object-level multi-modal adversaries is non-trivial due to three inherent constraints. First, the cross-modal signal gap: cameras and LiDAR capture fundamentally different signals (texture vs. geometry), making it difficult to find a single perturbation that simultaneously degrades both. Second, the dynamic viewing condition: autonomous driving involves drastic changes in distance and viewpoint; an adversary optimized for one pose often fails as the vehicle moves. Third, the black-box nature of production-grade stacks: real systems often involve non-differentiable preprocessing (e.g., NMS) and hide their internal gradients, necessitating a query-efficient optimization paradigm.

Most prior adversarial research targets a single sensing stream---either camera-only~\cite{huang2023t,wei2022simultaneously,wei2023unified,zhang2023capatch,bayer2026higher,huang2026advreal} or LiDAR-only~\cite{shin2017illusion,cao2019adversarial,sun2020towards,jin2023pla,cao2023you,sato2023lidar,zheng2025new}---and often assumes white-box access (or differentiable surrogates) to optimize perturbations. Recent 3D object attacks~\cite{lee2020shapeadv,cao2021invisible} further explore mesh- or shape-level manipulations, yet they typically rely on gradient-based optimization through renderers or model internals and are mainly evaluated against \emph{single-modality} detectors; as a result, the optimized geometries may deviate from natural object priors or overfit to limited viewpoints. These limitations become particularly pronounced in realistic driving scenarios with large viewpoint/distance variations, and they do not naturally account for the cross-modal alignment and consistency mechanisms inherent to Camera--LiDAR fusion stacks. This gap motivates an object-level, black-box attack paradigm that can synthesize a \emph{single} physically plausible 3D entity with controllable semantics and appearance, and optimize it to degrade both monocular and fusion detectors simultaneously. Crucially, such optimization must preserve physical plausibility and multi-view consistency, producing a stable adversary rather than a viewpoint-specific artifact.

Building on these insights, we propose \textbf{Agent-OBJ}, a prompt-driven generative framework that synthesizes physically plausible 3D adversaries and induces consistent cross-system failures in autonomous driving perception (Fig.~\ref{fig:1}).  As illustrated, Agent-OBJ instantiates a canonical pedestrian geometry from a base prompt, modulates its shape via an additional semantic prompt, and controls appearance using a pretrained multi-view personalization model conditioned on multi-view images and a style prompt to produce view-consistent textures. 
It then performs score-based, gradient-free optimization in a strict black-box setting, using only joint confidence feedback from a monocular 2D detector and a Camera--LiDAR fusion 3D detector. To ensure stealthiness, we regularize the generated point cloud by encouraging similarity to a pedestrian feature bank, aligning it with the distribution of natural objects. Moreover, we incorporate a multi-view consistency constraint to promote cross-view robustness under diverse viewpoints. Extensive evaluations on nuScenes demonstrate that Agent-OBJ achieves strong and stable adversarial effects against both monocular and fusion detectors while remaining highly inconspicuous.

Our main contributions are four-fold:
\begin{itemize}
    \item We formulate a practical \emph{black-box, object-level} threat model for \emph{unified} attacks on monocular 2D and Camera--LiDAR fusion 3D perception, emphasizing multi-view robustness under realistic driving viewpoints.
    \item We propose Agent-OBJ, a prompt-driven generative pipeline that decouples \emph{semantic} (geometry) and \emph{style} (appearance) control, enabling discrete, controllable exploration of physically plausible 3D adversarial objects.
    \item We introduce a joint optimization objective with naturalness and multi-view stability regularization, producing adversaries that are both effective across modalities and visually/geometrically plausible.
    \item We demonstrate strong attack success, transferability across detectors, and improved query efficiency on nuScenes under a fixed black-box query budget.
\end{itemize}

\section{Related Work}
\label{related}

\noindent\textbf{Camera-based Attacks.}
Attacks on monocular perception typically rely on \emph{pixel-level} perturbations \cite{xie2017adversarial,wei2018transferable} or localized \emph{adversarial patches} \cite{doan2022tnt,brown2017adversarial}. Early physical-world studies highlight the importance of \emph{transformation robustness} by optimizing over viewpoint, scale, illumination, and printing effects (e.g., EOT/RP2) \cite{athalye2018synthesizing,eykholt2018robust,song2018physical}. Recent works further improve patch deployability for detection via triggerable/camouflaged designs and robustness-oriented training or decomposition strategies \cite{zhu2023tpatch,cheng2024depatch,huang2023t,wei2022simultaneously,wei2023unified,zhang2023capatch,bayer2026higher,huang2026advreal}. Beyond detection, adversarial data augmentation has also been explored to improve robustness in low-data regimes such as few-shot classification \cite{xu2025noise}.

Despite strong progress, these methods remain fundamentally constrained to \emph{image-space} manipulations: they primarily compromise the camera branch and often exhibit limited effectiveness as \emph{object-level} adversaries that generalize to multi-modal perception stacks. Beyond detection, related lines also study attacks on tracking and segmentation, but they similarly operate on camera-specific representations and typically do not address cross-modal consistency required by modern fusion systems.

\noindent\textbf{LiDAR-based 3D Attacks.}
Threats to 3D perception span \emph{digital} point-cloud manipulations \cite{wang2021adversarial,wang2023adversarial,liu2023slowlidar} and \emph{physical} interventions.
The latter includes active sensor spoofing (e.g., laser injection and aiming) \cite{shin2017illusion,cao2019adversarial,sun2020towards,jin2023pla,cao2023you,sato2023lidar,suzuki2024wip},
passive spoofing via reflective surfaces \cite{kobayashi2025invisible}, and other environmental exploits such as shadow/smoke \cite{zheng2025new,kobayashi2025invisible}.
Beyond point-wise noise, \emph{object-level} physical attacks seek adversarial 3D structures that persist across viewpoints and distances \cite{cao1907adversarial,yang2021robust,zhu2021can}, and recent work emphasizes reproducible evaluation and sim-to-real transfer \cite{cheng2025revisiting}.

A recurring challenge in LiDAR attacks is the gap between simulation-time perturbations and real sensors, where beam patterns, range-dependent noise, and occlusion can substantially change point returns. This motivates both stronger physical modeling and robustness-oriented evaluation protocols, especially when the attacker must remain effective under realistic ego-motion and viewpoint changes.

\noindent\textbf{Fusion Attacks and Object-level Adversaries.}
Multi-sensor fusion is designed to suppress modality-specific noise via cross-modal cues, making unified attacks harder than single-sensor ones. However, existing fusion attacks are still limited: many rely on unrealistic white-box access or perturb only one branch, while black-box attempts (e.g., frustum attack~\cite{hallyburton2022security}) mainly target specific semantic-level fusion assumptions. Meanwhile, recent evidence~\cite{huang2024comparison,su2020robustfusion} shows that even single-modality perturbations can spill over to fusion detectors, implying that the fusion stack exposes non-trivial attack surfaces beyond individual sensors. From a system viewpoint, failures often arise from joint decision logic—misalignment between modalities, confidence domination of one branch, or error amplification in association/NMS—yet most attacks do not model or optimize against these coupling mechanisms. Finally, differentiable-rendering 3D adversaries~\cite{athalye2018synthesizing,xiao2019meshadv,zeng2019adversarial,oslund2022multiview,suryanto2022dta,suryanto2023active,wang2022fca} are predominantly texture-centric over predefined meshes, typically tied to digital rendering, and rarely validated for LiDAR or for cross-system robustness under realistic spatial variations. 

These gaps motivate attacks that are black-box, object-level, and explicitly account for fusion coupling rather than optimizing a single modality in isolation.

\noindent\textbf{Neural Rendering and Generative 3D Adversaries.}
With the rise of neural scene/object representations, recent work\cite{li2024adv3d,meng2025gaussian,zhang20253dgaa,lou20253d} explores adversarial objects via NeRF and 3D Gaussian Splatting (3DGS), enabling efficient multi-view optimization and rapid rendering. In parallel, generative 3D pipelines (text-to-3D, text-to-multi-view, and multi-view personalization) provide a natural mechanism to explore semantic variations beyond continuous texture noise, offering higher-level controllability and potentially better physical plausibility. Such controllable generative editing has also been systematically benchmarked for fine-grained attribute manipulation (e.g., drag-based gaze direction transformation)~\cite{fu2025benchmarking}. However, many existing neural-rendering attacks still rely on differentiable objectives and continuous parameter tuning, which may not match strict query-only settings and can overfit to limited viewpoints. 

In contrast, we study a practical \emph{object-level} threat model under strict \emph{black-box} constraints, and search a discrete, text-conditioned generation space to induce meaningful geometry and appearance variations. Guided solely by confidence feedback from monocular and Camera--LiDAR fusion detectors, our approach yields physically plausible adversarial objects that transfer across perception pipelines and remain stable under diverse viewpoints and realistic driving conditions.

\section{Method}
\label{sec:method}

\begin{figure*}[t] 
\centerline{\includegraphics[width=\linewidth]{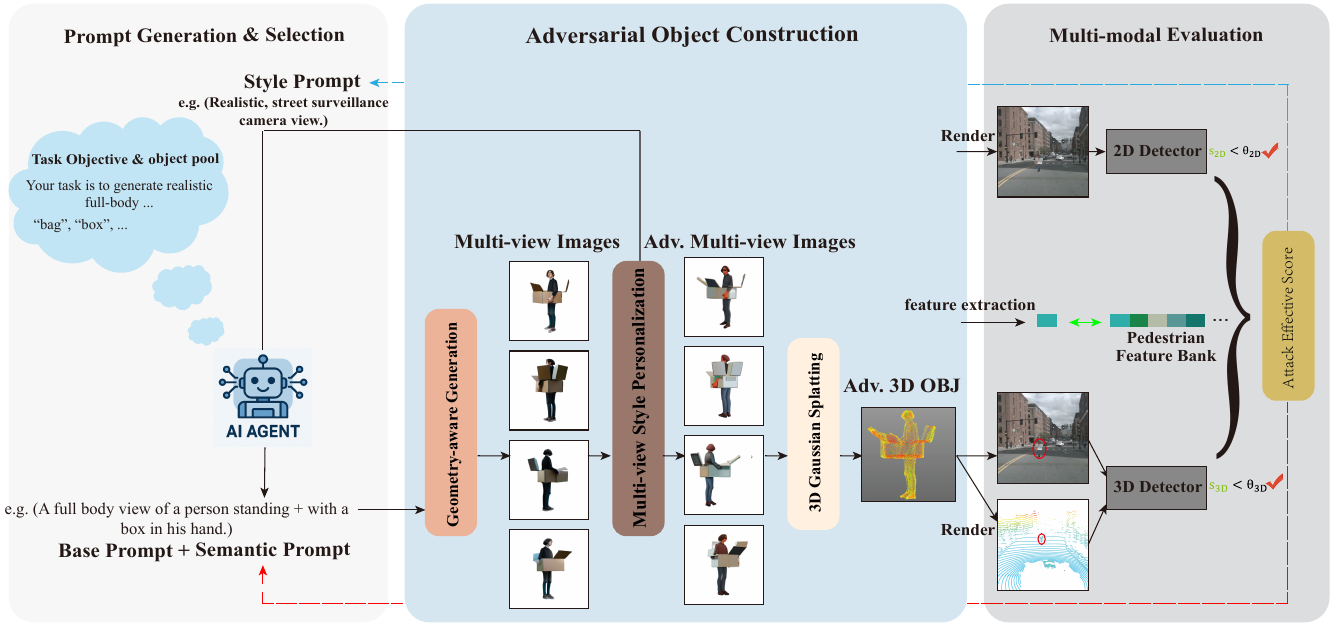}} 
\caption{The architecture of Agent-OBJ. This three-stage pipeline leverages an AI Agent for prompt steering, 3D Gaussian Splatting for adversarial object construction, and a multi-modal evaluation suite to evaluate disappearance attacks against state-of-the-art detectors.} 
\label{fig:2} 
\end{figure*}

\paragraph{Overview.}
Fig.~\ref{fig:2} illustrates the overall three-stage workflow of Agent-OBJ. Starting from a base prompt $p^{base}$, an LLM agent iteratively edits the semantic/style prompts $(p^{sem},p^{sty})$ to instantiate a candidate pedestrian-like 3D object (Sec.~\ref{sec:generation}). The object is then rendered and inserted into calibrated camera--LiDAR observations via $\mathcal{R}$ and evaluated by a monocular 2D detector $f_{2D}$ and a fusion 3D detector $f_F$ over viewpoints $\mathcal{V}$ (Sec.~\ref{sec:threat_model}). The resulting confidence feedback, together with naturalness and multi-view stability regularizers, defines a joint objective that guides prompt updates under a fixed query budget $B$ (Sec.~\ref{sec:optimization}).

\subsection{Problem Definition and Threat Model}
\label{sec:threat_model}

We study a practical black-box, object-level attack on an autonomous-driving perception stack comprising a monocular 2D detector $f_{2D}$ and a Camera--LiDAR fusion 3D detector $f_F$. The adversary physically inserts a single pedestrian-like object $\mathcal{O}$ at a fixed location in front of the ego vehicle and seeks targeted disappearance: $\mathcal{O}$ should be missed by both detectors over a set of realistic viewpoints $v\in\mathcal{V}$, instantiated by rotating the object’s yaw relative to the ego vehicle. The attacker is query-only: each query exposes only post-processed detections (bounding boxes and confidence scores), with no access to logits, gradients, intermediate features, or training data, and is constrained by a query budget $B$.

Given a scene $\mathcal{S}$ and viewpoint $v$, rendering/insertion yields synchronized multi-sensor observations
$(I_v,P_v)=\mathcal{R}(\mathcal{S},\mathcal{O},v)$, where $I_v$ denotes the rendered camera images and $P_v$ denotes the rendered LiDAR point cloud.

\paragraph{Insertion and rendering operator.}
We implement $\mathcal{R}$ as a calibration-consistent insertion operator that produces both camera and LiDAR observations. For cameras, $\mathcal{R}$ renders $\mathcal{O}$ into the target view using known extrinsics/intrinsics and alpha-composites it onto the background image while preserving depth ordering. For LiDAR, $\mathcal{R}$ places $\mathcal{O}$ in the ego frame and simulates its point returns using a beam-based forward model with range-dependent noise; the simulated returns are merged into the original scene point cloud. This design ensures each query reflects the detectors' \emph{final} post-processed predictions under a unified multi-sensor observation.

For detector $m\in\{2D,F\}$, we define the target confidence at $v$ as the maximum target-class confidence among detections overlapping the object's support
$\mathcal{B}_v^{\mathcal{O}}$:
\begin{equation}
s_m(v) = \max_{b\in \mathcal{D}_m(\cdot)} \left\{ \mathrm{conf}_{\text{target}}(b)\ \big|\ \mathrm{IoU}(b,\mathcal{B}_v^{\mathcal{O}})>\tau_{\mathrm{IoU}} \right\},
\end{equation}
where $\mathcal{D}_m(\cdot)$ denotes the \emph{final post-NMS} detections of $f_m$ and we set $s_m(v)=0$ if no box satisfies the overlap criterion.

\paragraph{Object support and matching criterion.}
The support $\mathcal{B}_v^{\mathcal{O}}$ is determined by the known insertion transform of $\mathcal{O}$. For $f_{2D}$, we obtain $\mathcal{B}_v^{\mathcal{O}}$ by projecting the inserted 3D bounding box onto the image plane and taking its tight 2D enclosing rectangle. For $f_F$, $\mathcal{B}_v^{\mathcal{O}}$ is the inserted 3D bounding box in the LiDAR/ego frame, and we match predictions using \emph{3D IoU} under the detector’s box parameterization. We apply an IoU gate $\tau_{\mathrm{IoU}}$ to associate predicted boxes with the inserted object; if no post-NMS prediction overlaps $\mathcal{B}_v^{\mathcal{O}}$, we set $s_m(v)=0$. The attack succeeds when both $s_{2D}(v)$ and $s_F(v)$ are suppressed below their respective decision thresholds.

\paragraph{Placement and viewpoints.}
We insert $\mathcal{O}$ at a fixed range along the ego vehicle's forward axis so that it is visible in the front camera, and evaluate robustness under a discrete viewpoint set $\mathcal{V}$ realized by rotating the object's yaw relative to the ego vehicle.
In our default setting, $|\mathcal{V}|=4$ (yaw angles specified in Sec.~\ref{sec:exp_setup}).

\paragraph{Decision thresholds and query accounting.}
We treat each detector as a black-box decision system with a fixed post-processing threshold $\theta_m$ (confidence and NMS).
A viewpoint is considered \emph{missed} if $s_m(v) < \theta_m$ or the overlapping detection is suppressed by NMS.
Each evaluation of one object queries both detectors over all viewpoints, costing $2|\mathcal{V}|$ queries.
Thus, $T$ optimization iterations correspond to a total query budget of $B = 2T|\mathcal{V}|$ queries (or $T|\mathcal{V}|$ per detector).

\paragraph{Viewpoint-robust success criterion.}
We optimize for joint failures across $v\in\mathcal{V}$ rather than a single ``lucky'' pose. To quantify viewpoint robustness, we report the mean performance over $\mathcal{V}$ together with its cross-view variation, reflecting both effectiveness and stability.

\subsection{Prompt-conditioned 3D Adversarial Object Generation}
\label{sec:generation}

\paragraph{Decoupled prompts.}
We parameterize object generation with a base prompt $p^{base}$ (e.g., ``a full-body view of a person standing''),
a \emph{semantic} prompt $p^{sem}$ that controls geometry/shape (e.g., ``with a bag in hand''),
and a \emph{style} prompt $p^{sty}$ that controls appearance (e.g., ``realistic, street surveillance camera view'').
This decoupling allows us to adjust shape and appearance with controllable degrees of freedom during optimization.

\paragraph{Why decoupling helps black-box optimization.}
In a strict query-only setting, directly optimizing continuous 3D parameters is prohibitively expensive.
We instead search over a structured, text-conditioned space where edits to $p^{sem}$ induce semantic/shape changes, while edits to $p^{sty}$ primarily affect appearance.
This factorization yields interpretable, low-entropy updates, improving query efficiency over unconstrained perturbations.

\paragraph{Multi-view synthesis and personalization.}
A text-to-multi-view generator produces view-consistent images for geometry initialization:

\begin{equation}
\mathcal{I}_{geo}=G_{\mathrm{T2MV}}\!\left([p^{base};p^{sem}]\right),
\end{equation}
where $\mathcal{I}_{geo}$ contains a set of multi-view images with consistent structure. We then apply a pretrained multi-view personalization module, conditioned on multi-view images and $p^{sty}$, to obtain view-consistent adversarial appearance:
\begin{equation}
\mathcal{I}_{adv}=\Psi_{\mathrm{edit}}(\mathcal{I}_{geo},p^{sty}),
\end{equation}
which edits texture/color/material while preserving multi-view consistency.

\paragraph{Two-stage multi-view consistency.}
We enforce view consistency twice: (i) $G_{\mathrm{T2MV}}$ stabilizes geometry by producing coherent multi-view structure, and (ii) $\Psi_{\mathrm{edit}}$ applies appearance edits conditioned on the multi-view set to prevent view-specific textures that break under yaw changes.
This is critical because we evaluate and regularize performance over multiple viewpoints.

\paragraph{3D instantiation via 3DGS.}
Finally, we reconstruct a 3D Gaussian Splatting object from $\mathcal{I}_{adv}$:
\begin{equation}
\mathcal{O}=\mathrm{3DGS}(\mathcal{I}_{adv}),
\end{equation}
yielding a renderable object that supports photorealistic camera rendering and LiDAR simulation within $\mathcal{R}$.

\paragraph{3DGS representation.}
3DGS represents $\mathcal{O}$ as a set of anisotropic Gaussians with parameters
$\{ \boldsymbol{\mu}_i,\boldsymbol{\Sigma}_i,\alpha_i,\mathbf{c}_i(\cdot)\}_{i=1}^N$,
where $\boldsymbol{\mu}_i$ and $\boldsymbol{\Sigma}_i$ denote center and covariance, $\alpha_i$ is opacity, and $\mathbf{c}_i$ denotes view-dependent radiance.
This representation supports fast differentiable rendering for cameras and efficient point sampling for LiDAR simulation, enabling repeated evaluation within the query budget.

\paragraph{Representation and controllable factors.}
In our prompt-conditioned pipeline, $p^{sem}$ primarily drives \emph{shape/structure} (e.g., body silhouette and carried items) through $G_{\mathrm{T2MV}}$,
while $p^{sty}$ mainly affects \emph{appearance} (e.g., clothing texture, color, material) through $\Psi_{\mathrm{edit}}$.
This separation enables targeted edits during optimization: semantic edits explore geometry-level variations, whereas style edits refine appearance while maintaining multi-view coherence.

\subsection{Joint Black-box Optimization with Stealthiness and Multi-view Stability}
\label{sec:optimization}

\paragraph{Threshold-aware attack objective.}
Directly minimizing raw confidence can be suboptimal because detectors make decisions by thresholding.
We therefore adopt a threshold-aware surrogate:
\begin{equation}
\phi_m(s) = \max(s-\theta_m,\,0),
\end{equation}
which penalizes confidence only when it exceeds the detector's decision threshold $\theta_m$.
Our per-view attack loss is then
\begin{equation}
\begin{aligned}
\mathcal{L}_{atk}(\mathcal{O};v)
&= \lambda_{2D}\,\phi_{2D}\!\left(s_{2D}(v)\right)
 + \lambda_F\,\phi_F\!\left(s_F(v)\right), \\
\mathcal{L}_{atk}(\mathcal{O})
&= \frac{1}{|\mathcal{V}|}\sum_{v\in\mathcal{V}}\mathcal{L}_{atk}(\mathcal{O};v).
\end{aligned}
\end{equation}

\paragraph{Joint aggregation across modalities.}
We optimize a unified objective that couples the two detectors at the score level rather than attacking each branch independently. In practice, we set $(\lambda_{2D},\lambda_F)$ to balance the two confidence scales and prevent one branch from dominating optimization. This joint formulation is essential for achieving high ASR$_{joint}$, as single-modality objectives tend to overfit to one detector and transfer poorly to the other.

\paragraph{Stealthiness and stability regularization.}
We optimize a full objective that balances attack strength, naturalness, and cross-view robustness:
\begin{equation} 
\mathcal{L}(\mathcal{O})=\mathcal{L}_{atk}(\mathcal{O})+\alpha\,\mathcal{L}_{nat}(\mathbf{X})+\beta\,\mathcal{L}_{mv}(\mathcal{O}),
\label{eq:full_obj_compact}
\end{equation}
where $\mathbf{X}$ denotes the object's geometry (point samples from 3DGS).

\paragraph{Naturalness regularizer $\mathcal{L}_{nat}$.}
To avoid unrealistic artifacts, we regularize the object to remain pedestrian-like in both geometry and semantics:
\begin{equation}
\mathcal{L}_{nat}(\mathbf{X}) =
\underbrace{\|\mathbf{g}(\mathbf{X})-\mathbf{g}_{ref}\|_1}_{\text{box-scale prior}}
+ \eta\,\underbrace{\mathrm{Dist}\!\left(\Phi(\mathbf{X}),\,\mathcal{M}_{ped}\right)}_{\text{feature plausibility}},
\end{equation}
where $\mathbf{g}(\mathbf{X})$ denotes the object's 3D box size statistics $(w,h,l)$ computed from the inserted object, and $\mathbf{g}_{ref}$ is the reference box-size statistic averaged over benign pedestrian instances. $\Phi(\mathbf{X})$ extracts a PointNet \cite{qi2017pointnet} embedding from the object point cloud, and $\mathcal{M}_{ped}$ is a pedestrian feature bank built by encoding nuScenes pedestrian point-cloud instances with the same PointNet. $\mathrm{Dist}(\cdot,\mathcal{M}_{ped})$ is implemented as the minimum cosine distance to the bank. 

\paragraph{Reference priors and feature bank.}
We use $\mathbf{g}(\mathbf{X})=(w,h,l)$ as the box-scale prior and compute $\mathbf{g}_{ref}$ as the mean box dimensions from benign pedestrian instances. To model semantic plausibility, we construct $\mathcal{M}_{ped}$ by extracting PointNet embeddings from nuScenes pedestrian point-cloud instances under the same preprocessing and encoder. We define $\mathrm{Dist}$ as the minimum cosine distance to $\mathcal{M}_{ped}$, which penalizes out-of-distribution artifacts and aligns optimization with the stealthiness metric used in experiments.

\paragraph{Multi-view stability regularizer $\mathcal{L}_{mv}$.}
To avoid view-specific attacks, we encourage consistent effectiveness across viewpoints by penalizing cross-view variation:
\begin{equation}
\mathcal{L}_{mv}(\mathcal{O})=\mathrm{Std}_{v\in\mathcal{V}}\!\left[\mathcal{L}_{atk}(\mathcal{O};v)\right],
\end{equation}
which promotes stable suppression across $v\in\mathcal{V}$ rather than overfitting to a single pose.

\paragraph{Budget-aware update schedule.}
We adopt a simple schedule to allocate the limited budget across semantic and style edits.
When the object becomes implausible (low naturalness), we prioritize semantic edits to restore structure, whereas when one detector remains confident, we adjust the complementary factor (shape or appearance) to reduce its score.
This scheduling reduces wasted queries compared to unguided random edits.

\paragraph{Prompt-driven agent search under black-box feedback.}
We perform gradient-free optimization by iteratively editing $(p^{sem},p^{sty})$ with an LLM agent under strict black-box feedback.
The agent operates in a constrained action space of discrete prompt edits, including \emph{add/replace/remove} of attribute phrases, while keeping $p^{base}$ fixed to preserve the pedestrian identity.
At iteration $t$, the agent proposes one edit to either $p^{sem}$ or $p^{sty}$ based on a compact history $\mathcal{H}_{t-1}$ that summarizes recent prompts and scalar feedback $\{s_{2D}(v),s_F(v)\}_{v\in\mathcal{V}}$ along with regularizer values.
We instantiate $\mathcal{O}_t$ via Sec.~\ref{sec:generation}, query both detectors over all $v\in\mathcal{V}$, and compute $\mathcal{L}(\mathcal{O}_t)$. We maintain the best-so-far object $\mathcal{O}^\star$; if $\mathcal{L}(\mathcal{O}_t) < \mathcal{L}(\mathcal{O}^\star)$, we accept the update and set $\mathcal{O}^\star\leftarrow\mathcal{O}_t$. Otherwise, we can keep the previous prompts (rollback) while recording the attempt for future edits. The procedure repeats until the query budget $B$ is exhausted, and we report $\mathcal{O}^\star$. Algorithm~\ref{alg:agentobj} summarizes the overall procedure. 

\begin{algorithm}[t]
\caption{Agent-OBJ: Black-box prompt-driven optimization}
\label{alg:agentobj}
\KwIn{Scene $\mathcal{S}$, base prompt $p^{base}$, initial prompts $(p^{sem}_0,p^{sty}_0)$, viewpoints $\mathcal{V}$, budget $B$}
\KwOut{Adversarial object $\mathcal{O}^\star$}
Initialize $\mathcal{O}^\star \leftarrow \mathrm{3DGS}(\Psi_{\mathrm{edit}}(G_{\mathrm{T2MV}}([p^{base};p^{sem}_0]),p^{sty}_0))$\;
Compute $\mathcal{L}(\mathcal{O}^\star)$ by querying $f_{2D},f_F$ over all $v\in\mathcal{V}$\;
Set number of iterations $T \leftarrow \lfloor B / (2|\mathcal{V}|) \rfloor - 1$ (the $-1$ accounts for the initial evaluation of $\mathcal{O}^\star$)\;
\For{$t=1$ \KwTo $T$}{
    $(p^{sem}_t,p^{sty}_t) \leftarrow \mathrm{EditAgent}(\mathcal{H}_{t-1})$\;
    Instantiate $\mathcal{O}_t$ via Sec.~\ref{sec:generation}\;
    Query both detectors over all $v\in\mathcal{V}$ and compute $\mathcal{L}(\mathcal{O}_t)$\;
    \eIf{$\mathcal{L}(\mathcal{O}_t) < \mathcal{L}(\mathcal{O}^\star)$}{
        \textbf{Accept} edit: $\mathcal{O}^\star \leftarrow \mathcal{O}_t$\;
    }{
        \textbf{Reject} edit: roll back to the previous prompts\;
    }
    Update history $\mathcal{H}_t$\;
    Stop if budget $B$ is exhausted\;
}
\Return $\mathcal{O}^\star$\;
\end{algorithm}

\paragraph{Complexity and budget.}
Each iteration evaluates one instantiated object by querying both detectors over all viewpoints, requiring $2|\mathcal{V}|$ black-box queries. Therefore, the total number of queries scales linearly with iterations, i.e., $B=2T|\mathcal{V}|$. The computational overhead outside detector queries is dominated by prompt-conditioned 3D instantiation (multi-view synthesis, personalization, and 3DGS reconstruction), which can be amortized by caching intermediate outputs when only $p^{sty}$ changes while $p^{sem}$ remains fixed.

\paragraph{Cost breakdown and caching.}
Each iteration incurs $2|\mathcal{V}|$ detector queries; thus the query complexity is $\mathcal{O}(T|\mathcal{V}|)$.
We cache intermediate results whenever possible: if only $p^{sty}$ changes, we reuse the geometry initialization from $G_{\mathrm{T2MV}}([p^{base};p^{sem}])$ and apply appearance edits on top, avoiding redundant generation. This amortization is beneficial under tight budgets where many iterations involve small appearance refinements.

\section{Experiments} 
\label{sec:experiments} 

\subsection{Experimental Setup} 
\label{sec:exp_setup}
\noindent\textbf{Datasets.} We evaluate on nuScenes~\cite{caesar2020nuscenes}, which contains 1,000 driving scenes (about 20 seconds each), around 40k annotated keyframes, roughly 1.4M camera images, and about 390k LiDAR sweeps. We use the official calibration and timestamps to generate synchronized multi-sensor observations.

\noindent\textbf{Attack protocol.} We select 10 scenes from the validation split and sample 5 keyframes per scene. For each keyframe, we insert a single pedestrian-like object at a fixed range (5--15\,m) along the ego vehicle's forward axis to ensure visibility in the front camera. This range covers the safety-critical zone where camera and LiDAR detection confidence peaks and the autonomous vehicle's reaction constraints are most stringent; beyond 15\,m, LiDAR sparsity already degrades all detectors independently of any attack. We evaluate robustness under four viewpoints $\mathcal{V}$ by rotating the object's yaw relative to the ego vehicle while keeping its position fixed. Unless otherwise stated, results are averaged over all scenes, keyframes, and viewpoints. 

\noindent\textbf{Detectors.} We consider a monocular 2D detector $f_{2D}$ (YOLOv10x~\cite{wang2024yolov10}) and a Camera--LiDAR fusion 3D detector $f_F$ (MambaFusion~\cite{wang2025mambafusion}). We use official checkpoints with default inference settings and standard post-processing (NMS). During optimization, we use Qwen2.5~\cite{qwen2025qwen25} as the language agent to propose prompt edits, and query only the detectors' final outputs (bounding boxes and confidence scores) as black-box feedback.

\noindent\textbf{Query budget.} We optimize one adversarial object \emph{per keyframe}. Each iteration evaluates one candidate by querying both detectors over all viewpoints in $\mathcal{V}$, costing $2|\mathcal{V}|$ queries. We run $T=100$ iterations, resulting in a per-keyframe budget of $B=2T|\mathcal{V}|=800$ total queries (i.e., $4T$ per detector for $|\mathcal{V}|=4$), and report the best-so-far object within the budget. Results are averaged over all keyframes.

\subsection{Evaluation Metrics} 
\label{sec:metrics} 
We report Attack Success Rate (ASR, \%) as the proportion of trials in which the inserted target is missed. Specifically, ASR$_{2D}$ and ASR$_F$ are computed w.r.t.\ the monocular and fusion detector, respectively,
and ASR$_{joint}$ requires both detectors to miss the target. We additionally report \textbf{confidence drop} as the reduction in the maximum target-class confidence before and after attack, denoted by $\Delta s_{2D}$ and $\Delta s_F$. To quantify \textbf{stealthiness}, we measure (i) deviation of geometry statistics from benign references and (ii) feature distance to natural pedestrians (lower is better). Finally, we evaluate \textbf{multi-view stability} over $\mathcal{V}$ by reporting the mean ASR across viewpoints together with its cross-view variation (Std/Var), or worst-view ASR when applicable. 

\noindent\textbf{Implementation details.} 
We instantiate a 3DGS object by reconstructing from four view-consistent images using a pretrained LGM~\cite{tang2024lgm}. The intermediate multi-view images are appearance-personalized with MV-Adapter~\cite{huang2025mv} conditioned on the style prompt $p^{sty}$, producing view-consistent texture/color/material edits at azimuths $0^\circ, 90^\circ, 180^\circ$, and $270^\circ$; low-opacity Gaussians are filtered out. We evaluate robustness on the same four canonical viewpoints $\mathcal{V}=\{0^\circ, 90^\circ, 180^\circ, 270^\circ\}$, implemented as object-yaw rotations in the ego frame. We use an IoU gate $\tau_{\mathrm{IoU}}=0.5$ for association (2D projection for $f_{2D}$ and 3D IoU in the LiDAR/ego frame for $f_F$). LiDAR returns are simulated via beam-based ray casting with distance-dependent range noise $\sigma(r)=\sigma_0+kr$ using a 32-beam pattern, with $\sigma_0=0.02$ and $k=10^{-4}$ (in meters), and are merged into the scene point cloud for MMDetection3D inference. We build the pedestrian feature bank $\mathcal{M}_{ped}$ by encoding 71,073 nuScenes pedestrian instances with PointNet~\cite{qi2017pointnet} and define $\mathrm{Dist}$ as the minimum cosine distance to the bank. A query corresponds to running one detector at one viewpoint, hence the total budget is $B=2T|\mathcal{V}|$. Unless stated otherwise, we use $\lambda_{2D}=\lambda_F=1$, $\alpha=\beta=0.1$, and $\eta=1$. The EditAgent draws \emph{add}/\emph{replace}/\emph{remove} operations from two disjoint pools, $\mathcal{D}_{sem}$ (geometry) and $\mathcal{D}_{sty}$ (texture), each containing ${\sim}30$ candidate attribute phrases sampled independently, and retains only loss-reducing edits. We will release the full prompts, edit lists, and code to ensure reproducibility.


\subsection{Main Results}
\label{sec:main_results}
As shown in Tab.~\ref{tab:main_results}, we report the main results on nuScenes. The Benign baseline inserts a normally generated pedestrian object without adversarial optimization. RandEdit removes agent guidance and instead applies random prompt edits from a predefined edit set, generating a different candidate object at each iteration under the same query budget. 2D-only and Fusion-only optimize the object using feedback from only the monocular detector or only the fusion detector, respectively.

Agent-OBJ achieves strong effectiveness on both branches (ASR$_{2D}$=68.0 and ASR$_F$=96.0), leading to a substantially higher joint success rate (ASR$_{joint}$=64.0) than all baselines. Compared with RandEdit and single-modality objectives, Agent-OBJ better balances unified attackability with robustness: it attains the lowest cross-view variation (MV-Std=5.3) while keeping FeatureDist close to the benign reference (0.12 vs.\ 0.08), indicating improved stability and perceptual plausibility under the same protocol.

\begin{table*}[t]
\centering
\caption{Main results on nuScenes (10 val scenes). We report attack success rates (ASR), confidence drop ($\Delta s$; measured relative to the Benign Prompt-to-3D baseline), multi-view stability (Std over $|\mathcal{V}|=4$ viewpoints), and stealthiness. “$\pm$” denotes standard deviation over keyframes.}
\label{tab:main_results}
\resizebox{\textwidth}{!}{%
\begin{tabular}{c|ccc|cc|cc}
\toprule
\multirow{2}{*}{Method} &
\multicolumn{3}{c|}{Attack Effectiveness $\uparrow$} &
\multicolumn{2}{c|}{Confidence Drop $\uparrow$} &
\multicolumn{2}{c}{Stealth \& Stability $\downarrow$} \\
& ASR$_{2D}$ & ASR$_F$ & ASR$_{joint}$ & $\Delta s_{2D}$ & $\Delta s_F$ & FeatureDist & MV-Std \\
\midrule
Benign (Prompt-to-3D)              & 5.0$\pm$2.1& 20.0$\pm$1.6 & 1.0$\pm$0.7 & -- & -- & 0.08 & 2.7 \\
RandEdit (query-matched)           & 14.1$\pm$4.3 & 35.0$\pm$8.7& 13.9$\pm$1.2 & 0.16 & 0.28 & 0.35 & 22.5  \\
2D-only objective                  & 78.0$\pm$8.1 & 23.0$\pm$3.0 & 20.3$\pm$5.9 & 0.82 & 0.19 & 0.17 & 18.3 \\
Fusion-only objective              & 15.0$\pm$6.2 & 87.0$\pm$3.5 & 13.7$\pm$4.8 & 0.44 & 0.34 & 0.26 & 14.5 \\
\hline
\textbf{Agent-OBJ}                & 68.0$\pm$8.3 & 96.0$\pm$3.7 & 64.0$\pm$6.2 & 0.62 & 0.41 & 0.12 & 5.3 \\
\bottomrule
\end{tabular}%
}
\end{table*}

\begin{table}[ht]
\centering
\caption{Comparison with representative prior attacks. Image-only attacks (NI-FGSM, T-SEA, FCA, PGA) cannot report ASR$_F$/ASR$_{joint}$, and LiDAR-only attacks (AdvPC, ScAR) report ASR$_F$ only; ``--'' marks inapplicable entries. Several baselines require white-box access or operate in a different threat model, thus are not directly comparable to our strict black-box setting. Agent-OBJ attains the highest ASR$_{joint}$ despite the black-box, no-gradient constraint.}
\label{tab:baseline_compare}
\setlength{\tabcolsep}{4pt}
\begin{tabular}{llccc}
\toprule
Method & Threat Model & ASR$_{2D}$ & ASR$_F$ & ASR$_{joint}$ \\
\midrule
NI-FGSM~\cite{lin2019nesterov}                & White-box, Img-Det & 71.4 & --   & --   \\
T-SEA~\cite{huang2023t}                       & White-box, Img-Det & 57.2 & --   & --   \\
FCA~\cite{wang2022fca}                        & White-box, Img-Det & 58.4 & --   & --   \\
PGA~\cite{lou20253d}                          & White-box, Img-Det & 61.3 & --   & --   \\
\midrule
AdvPC~\cite{hamdi2020advpc}                   & White-box, LiDAR   & --   & 67.3 & --   \\
ScAR~\cite{lu2023scar}                        & White-box, LiDAR   & 12.5 & 49.8 & 16.3 \\
\midrule
Invisible Attack~\cite{cao2021invisible}      & White-box, Fusion  & 18.6 & 44.2 & 20.9 \\
Frustum Attack~\cite{hallyburton2022security} & Black-box, Fusion  & --   & 87.3 & 26.1 \\
\midrule
\textbf{Agent-OBJ (ours)}                     & \textbf{Black-box, Fusion} & \textbf{68.0} & \textbf{96.0} & \textbf{64.0} \\
\bottomrule
\end{tabular}
\end{table}

\subsection{Comparison with Prior Attacks}
\label{sec:prior_compare}
Tab.~\ref{tab:baseline_compare} reports representative prior attacks as \emph{reference baselines}.
These methods are \emph{not directly comparable} to Agent-OBJ, since some assume white-box access and/or different threat models; inapplicable entries are marked as ``--''. Image-space attacks (e.g., NI-FGSM) apply only to the camera branch, so we report ASR$_{2D}$ only, while LiDAR/3D attacks (e.g., AdvPC) are reported in ASR$_F$ only.
Existing physical/object-level baselines (Invisible Attack and ScAR) show limited \emph{simultaneous} failures, whereas Agent-OBJ achieves substantially higher unified impact (ASR$_{joint}=64.0$) under a strict black-box, query-only setting.

\subsection{Ablation Studies}
\label{sec:ablation}
Tab.~\ref{tab:ablation} ablates major components of Agent-OBJ and highlights their complementary roles.
Removing the joint objective substantially reduces ASR$_{joint}$ and increases cross-view variation, indicating that explicitly coupling the two detectors is critical for consistent unified failures. Disabling personalization further degrades ASR$_{joint}$ to 38.4 and increases FeatureDist, showing that view-consistent appearance control is important for both effectiveness and plausibility.

The naturalness term $\mathcal{L}_{nat}$ reveals a clear effectiveness--stealth trade-off: dropping it boosts ASR$_{joint}$ but significantly harms stealthiness and stability, with FeatureDist worsening and MV-Std increasing. Finally, removing the multi-view stability regularizer $\mathcal{L}_{mv}$ leads to the lowest ASR$_{joint}$ (30.5) and higher variance, confirming that enforcing cross-view consistency is necessary to avoid viewpoint-specific ``lucky'' attacks.

\begin{table}[ht]
\centering
\caption{Ablation studies on nuScenes. Each component is removed from Agent-OBJ.}
\label{tab:ablation}

\begin{tabular}{lccc}
\toprule
Variant & ASR$_{joint}\uparrow$ & FeatureDist$\downarrow$ & MV-Std$\downarrow$ \\
\midrule
Full (Agent-OBJ)               & 64.0 & 0.12 & 5.3 \\
w/o joint objective            & 40.8 & 0.11 & 9.4 \\
w/o personalization            & 38.4 & 0.15 & 8.2 \\
w/o $\mathcal{L}_{nat}$        & 80.4 & 0.36 & 16.8    \\
w/o $\mathcal{L}_{mv}$         & 30.5 & 0.14 & 8.3    \\
\bottomrule
\end{tabular}
\end{table}

\subsection{Transferability Across Detectors}
\label{sec:transfer}
Tab.~\ref{tab:transfer_simple2} evaluates transferability without re-optimization. When swapping the 2D detector (YOLOv10x$\rightarrow$DINO\cite{zhang2022dino}) while keeping the fusion model fixed, the adversarial object still achieves strong ASR$_F$ (96.0) and non-trivial joint success (53.7), suggesting that the learned adversary is not overfitted to a single monocular model. When swapping the fusion detector (MambaFusion$\rightarrow$BEVFusion \cite{liang2022bevfusion}) with the same 2D detector, ASR$_{joint}$ further improves to 66.2, indicating that the generated object generalizes across different fusion architectures. Beyond cross-model transfer, the object also generalizes \emph{across datasets}: applied zero-shot to KITTI~\cite{geiger2012kitti}, which differs in sensor configuration and calibration, it retains 58.9 ASR$_{joint}$, confirming that the attack exploits cross-modal fusion coupling rather than dataset-specific artifacts.

\begin{table}[t]
\centering
\caption{Transferability of the adversarial object optimized on (YOLOv10x, MambaFusion) in a black-box manner, reported \emph{without re-optimization}. The last row additionally transfers the object \emph{zero-shot} to a different dataset (KITTI).}
\label{tab:transfer_simple2}

\begin{tabular}{lccc}
\toprule
Test setting & ASR$_{2D}$ & ASR$_F$ & ASR$_{joint}$ \\
\midrule
In-domain: (YOLOv10x, MambaFusion)     & 68.0  & 96.0  & 64.0  \\
2D transfer: (DINO, MambaFusion)       & 57.6 & 96.0 & 53.7 \\
Fusion transfer: (YOLOv10x, BEVFusion) & 68.0 & 98.7 & 66.2 \\
\midrule
Cross-dataset: $\rightarrow$ KITTI~\cite{geiger2012kitti} (zero-shot) & 62.6 & 90.8 & 58.9 \\
\bottomrule
\end{tabular}
\end{table}

\subsection{Query Efficiency}
\label{sec:query}
As shown in Tab.~\ref{tab:query}, Agent-OBJ achieves the best query efficiency under the same black-box protocol. RandEdit applies unguided random edits and does not use the returned scores to steer subsequent updates, so many queries are spent on low-yield directions, resulting in slower gains and lower ASR$_{joint}$. Greedy is more efficient by selecting the locally best score-improving edit, but its myopic nature can get trapped in local optima and saturate earlier as the budget grows. In contrast, Agent-OBJ uses score feedback to guide more informed exploration and higher-quality edits, making each query more likely to translate into progress; this is consistent with its lower Queries@80\%, indicating faster convergence under a fixed budget.

\begin{table}[ht]
\centering
\caption{Query efficiency on nuScenes. ASR$_{joint}$ under different query budgets (total detector calls over both detectors and all $|\mathcal{V}|$ viewpoints). Queries@80\%: queries to reach 80\% of the final ASR$_{joint}$ at 800 queries.}
\label{tab:query}
\begin{tabular}{lcccc}
\toprule
Method & ASR$_{joint}$@200 & ASR$_{joint}$@400 & ASR$_{joint}$@800 & Queries@80\% $\downarrow$ \\
\midrule
RandEdit            & 2.8 & 6.7 & 13.9 & 712 \\
Greedy              & 31.5 & 37.0 & 61.2 & 648 \\
Agent-OBJ           & 37.0 & 45.9 & 64.0 & 584 \\
\bottomrule
\end{tabular}
\end{table}

\section{Conclusion}
Agent-OBJ exposes a critical blind spot in autonomous driving perception: a single, physically plausible 3D object can simultaneously degrade monocular and Camera–LiDAR fusion detectors, undermining the common redundancy-based safety assumption. At its core, Agent-OBJ employs an LLM agent to iteratively steer semantic and style prompts, searching a discrete text-conditioned generation space using only black-box confidence feedback from both detectors. Coupled with naturalness regularization and multi-view stability constraints, this agent-driven optimization produces stealthy and stable adversaries. Results on the nuScenes dataset highlight the urgency of developing stronger defenses against object-level, cross-modal attacks.

{
    \small
    \bibliographystyle{splncs04}
    \bibliography{ref}
}

\end{document}

%% file: blwritepackage.tex
\usepackage{amsmath}
\usepackage{amssymb}
\usepackage{soul}
\setstcolor{red}
\sethlcolor{yellow}

\usepackage{multirow}
\usepackage{tabularx}
\usepackage{booktabs}
\usepackage{longtable}
\usepackage{makecell}

\usepackage{tikz}

\usepackage{dsfont}
\usepackage{xcolor}
\usepackage{color}
\usepackage[export]{adjustbox}
\usepackage{comment}

\usepackage[utf8]{inputenc} 
\usepackage{url}            
\usepackage{amsfonts}       
\usepackage{nicefrac}       
\usepackage{microtype}      
\usepackage{enumitem}
\usepackage{blindtext}
\usepackage{sidecap}
\usepackage{wrapfig}

\definecolor{citecolor}{RGB}{65,105,225}

\definecolor{dg}{rgb}{0,0.694,0.298}
\definecolor{purple}{rgb}{0.4,0.176,0.569}
\definecolor{royalblue}{RGB}{65,105,225}
\definecolor{olive}{RGB}{124,130,96}
\definecolor{anolive}{RGB}{85,107,47}
\definecolor{rosybrown}{RGB}{188,133,160}
\definecolor{saddlebrown}{RGB}{139,69,19}

\usepackage{pifont}

\makeatletter
\DeclareRobustCommand\onedot{\futurelet\@let@token\@onedot}
\def\@onedot{\ifx\@let@token.\else.\null\fi\xspace}

\makeatother